\documentclass[letterpaper, 10 pt, conference]{ieeeconf}  

\IEEEoverridecommandlockouts                              

\newcommand{\citep}[1]{\cite{#1}}
\newcommand{\citet}[1]{\cite{#1}}

\newcommand{\B}[1]{\textbf{\underline{#1}}}

\usepackage{amsmath} 
\usepackage{amssymb}
\usepackage{amsfonts}
\usepackage{dsfont}
\usepackage{multicol}
\usepackage[pdftex]{graphicx} 
\usepackage{graphicx}
\usepackage{times}
\usepackage{color, colortbl, xcolor}
\usepackage{epsfig} 
\usepackage{mathptmx} 
\usepackage{times} 
\usepackage{tikz}
\usetikzlibrary{fit, positioning, arrows, calc}
\usepackage{caption}
\usepackage{multirow}
\usepackage{rotating}
\usepackage{outlines}
\usepackage{verbatim}
\usepackage{hyperref}
\usepackage{floatrow}
\usepackage{tabularx}
\newcommand{\ORCAeps}[0]{Noisy ORCA}
\DeclareMathOperator*{\argmax}{arg\,max}

\usepackage{algorithm,algorithmic}
\usepackage{amsmath,bm}
\usepackage{caption}
\usepackage{subcaption}
\usepackage[noadjust]{cite}
\usepackage{soul}

\title{\LARGE \bf
Stranger Danger! Identifying and Avoiding Unpredictable Pedestrians in RL-based Social Robot Navigation\vspace{-3mm} 
}

\author{Sara Pohland$^{*1}$, Alvin Tan$^{*1}$, Prabal Dutta$^{1}$, and Claire Tomlin$^{1}$\vspace{-1mm}
\thanks{$^{*}$ Both authors contributed equally to this work.}
\thanks{$^{1}$ Dept. of EECS, UC Berkeley, Berkeley, CA, USA.}
}

\begin{document}

\maketitle
\thispagestyle{empty}
\pagestyle{empty}

\begin{abstract}

Reinforcement learning (RL) methods for social robot navigation show great success navigating robots through large crowds of people, but the performance of these learning-based methods tends to degrade in particularly challenging or unfamiliar situations due to the models' dependency on representative training data. To ensure human safety and comfort, it is critical that these algorithms handle uncommon cases appropriately, but the low frequency and wide diversity of such situations present a significant challenge for these data-driven methods. To overcome this challenge, we propose modifications to the learning process that encourage these RL policies to maintain additional caution in unfamiliar situations. Specifically, we improve the Socially Attentive Reinforcement Learning (SARL) policy by (1)~modifying the training process to systematically introduce deviations into a pedestrian model, (2)~updating the value network to estimate and utilize pedestrian-unpredictability features, and (3)~implementing a reward function to learn an effective response to pedestrian unpredictability. Compared to the original SARL policy, our modified policy maintains similar navigation times and path lengths, while reducing the number of collisions by 82\% and reducing the proportion of time spent in the pedestrians' personal space by up to 19 percentage points for the most difficult cases. We also describe how to apply these modifications to other RL policies and demonstrate that some key high-level behaviors of our approach transfer to a physical robot.
\end{abstract}
\vspace{-0.5mm}

\section{Introduction}
\label{sec:intro}

While robot navigation has been explored extensively, smooth integration of mobile robots into human-populated spaces is yet to be achieved. Robots that interact with people are expected to navigate in a way that is predictable and unobtrusive, maintaining both the safety and comfort of surrounding people \citep{rios-martinez_proxemics_2015}. The social robot navigation field is seeing a growing number of RL-based approaches that implicitly predict human motion and plan robot paths without explicit models of human behavior \citep{mavrogiannis_core_2021}. These RL-based approaches have achieved great success in enabling effective navigation around large crowds of people, outperforming traditional approaches \citep{chen_decentralized_2016, faust2018, chiang2019, chen_crowd-robot_2019}. However, the performance of RL policies is contingent on having representative training data, so these policies are sensitive to differences in pedestrian behavior seen during deployment versus training (a problem generally referred to as \textit{domain shift}). Domain shift is always a concern with learning-based methods but is of particular importance in social robot navigation because of the wide range of human behavior and the potential physical hazards and psychological risks associated with mobile robots operating in close proximity to humans \citep{salvini_safety_2022}. 

\begin{figure}[t]
    \centering
    \captionsetup{width=\columnwidth}
    \includegraphics[width=\columnwidth]{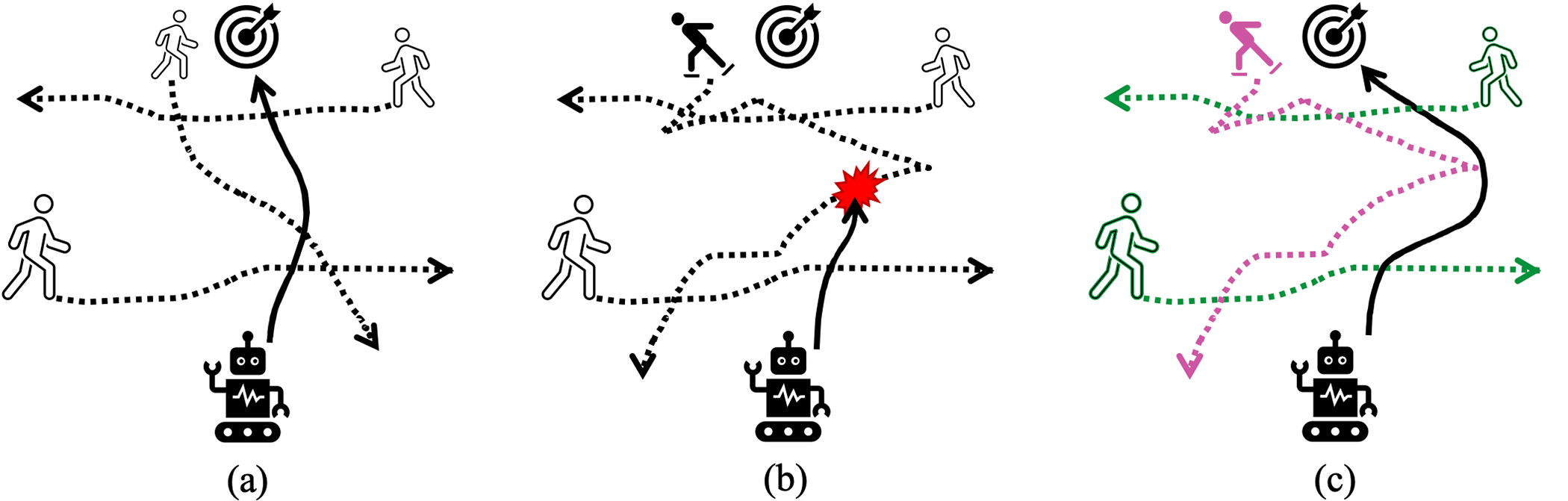}
    \caption{(a) RL-based robot navigation policies are trained with humans that behave according to some pedestrian model. (b) During deployment, these policies will encounter pedestrians that behave differently. Existing RL policies generally do not consider this and continue to treat all pedestrians the same, presenting concerns for human comfort and safety. (c) RL policies should distinguish between predictable (green) and unpredictable (pink) pedestrians and maintain appropriate caution while still navigating efficiently.\vspace{-5mm}}
    \label{fig:overview}
\end{figure}

Thus, to widely deploy RL-based approaches for social robot navigation, these methods should recognize their own level of \textit{uncertainty} in situations in which pedestrians behave \textit{unpredictably} from the perspective of the RL policy. We say that pedestrians behave unpredictably if their behavior deviates significantly from their normal or expected behavior as defined by the RL policy's implicit model of human behavior. Once an RL policy recognizes that it is in an unfamiliar situation and cannot accurately predict the behavior of nearby pedestrians, it should respond with appropriate caution (Figure~\ref{fig:overview}). Thus, it must \textit{distinguish} between pedestrians who exhibit predictable behavior seen during training and those whose behavior is unpredictable, and learn to navigate efficiently around predictable pedestrians while maintaining caution toward unpredictable ones. This would allow such policies to generalize well to arbitrary pedestrian behavior.

To explore this idea, we incorporate uncertainty-awareness into an existing RL policy called SARL~\citep{chen_crowd-robot_2019} by (1)~modifying the training process to systematically inject significant deviations into a model of pedestrian behavior, (2)~augmenting the observation space of the value network algorithm to recognize and quantify deviations of pedestrian behavior from the assumed model, and (3)~adding a term to the reward function to encourage caution toward progressively more unpredictable pedestrians while navigating normally around predictable ones. We conduct ablation studies in simulation to understand the cumulative impact of these modifications and find that they substantially improve the performance of SARL around particularly difficult and previously unseen pedestrian behavior. Compared to the original policy, our modified policy maintains similar navigation times and path lengths while notably reducing the number of collisions and the proportion of time spent in the simulated pedestrians' personal space. We then describe how the same modifications can be made to other socially-aware RL policies and demonstrate on hardware that our policy successfully identifies and maintains caution around real-world pedestrians who exhibit behaviors that are not part of the training distribution.\footnote{Code for reproducing our methods and analysis is available on GitHub: https://github.com/sarapohland/stranger-danger.}.

\section{Related Work}

We first note high-level procedural novelties of our work and then distinguish our approach from closely related work.

Traditional approaches to social robot navigation \textit{explicitly} predict human trajectories and then plan paths around them. Because these approaches use explicit human models, they can be adapted to detect when a pedestrian consistently deviates from these models, and then respond by calculating a conservative path that still maintains pedestrian comfort and safety~\cite{FisacBHFWTD18, bajcsy_scalable_2019, fridovich-keil_confidence-aware_2020, 9341469, 9210199, li_provably_2021, active, hu_sharp_2022}. Some RL-based policies also incorporate explicit pedestrian trajectory predictions \cite{multimodal_traj_predict, interaction_graph}, but the vast majority do not. Our work extends the anomaly detection and response process to RL-based policies that \textit{implicitly} model agent interactions and are thus not directly amenable to techniques designed for explicit human models.

The evaluation procedures in most prior work in RL-based social robot navigation are ill-suited for determining policy performance under significant deviations from the assumed pedestrian model (i.e., under domain shift) because (i) the evaluations are conducted using the same pedestrian model as was used during training, albeit in randomly generated scenarios and (ii) they usually only report average performance values, which reveal very little about policy performance in particularly difficult and unfamiliar situations~\citep{chen2018socially, sarl*, sathyamoorthy2020densecavoid, soadrl, chen2020relational, choi2021, gao_evaluation_2022}. To better evaluate and quantify policy performance under domain shift, we evaluate our policies on pedestrian models that are outside of the training distribution, and we report Conditional Value at Risk (CVaR) values, which describe expected performance on the hardest cases~\cite{risk}.




One approach to addressing the domain shift problem in RL-based social robot navigation could be to train on more realistic data (e.g., higher-fidelity pedestrian models or real-world pedestrian data) \cite{sean2, socialgym}. While this approach would expand what is included in the training distribution, there are undoubtedly myriad situations and behaviors that still lie outside the training distribution, so the need to identify and account for these unfamiliar situations still persists. In our paper, we attempt to express this gap in realism by training on a relatively simple and homogeneous pedestrian model and testing on scenarios that include a mix of three different pedestrian models with randomized parameters.

A collection of papers manage pedestrian unpredictability by training the robot to avoid regions around pedestrians called ``Danger Zones'' or ``Warning Zones'' that comprise all their physically plausible next states~\cite{danger_zones, warning_zone, Kstner2021EnhancingNS}. The size and shape of these Zones depend on pedestrian velocity and observed demographic (e.g., child vs. adult). In our approach, we directly adjust each pedestrian's discomfort distance instead of defining additional Zones, and these adjustments are based on inferred unpredictability values that describe the RL policy's training limits, rather than being directly related to the pedestrians' observed physical features.

One prior work quantifies pedestrian deviation from a given model, and then adaptively switches from a fast to slow RL policy when any pedestrian within a neighborhood of the robot is deemed unpredictable \cite{katyal_intent-aware_2020}. This approach forces the robot to respond either efficiently or cautiously towards all surrounding pedestrians, even if the majority of nearby pedestrians are acting predictably. In contrast, our approach identifies \textit{specific individuals} around which the robot should be more cautious and integrates this information directly into the RL policy, allowing the robot to exercise individualized caution as appropriate while still navigating efficiently around all other predictable pedestrians.

Our work builds on SARL~\citep{chen_crowd-robot_2019}, which is an RL-based method for crowd-aware robot navigation that predicts the optimal robot action given the current state of the robot and the configuration of the crowd. While SARL performs well around the types of pedestrians on which it was trained, it is highly dependent on its training data~\cite{sarl*}, generalizes poorly to novel situations~(as we demonstrate in Section~\ref{sec:results}), and has no way of recognizing when it is in an unfamiliar situation. 
\section{Our Uncertainty-Aware RL Policy} 
\label{sec:policy}

To reduce SARL's dependence on its training data, allow the policy to recognize when it is in an unfamiliar situation, and improve the ability of the policy to generalize to novel scenarios, we modify the training process~(\S \ref{subsec:training}), the model architecture~(\S \ref{subsec:model}), and the reward function~(\S \ref{subsec:reward}). 

\begin{figure}
    \centering
    \captionsetup{width=\columnwidth}
    \includegraphics[width=\columnwidth]{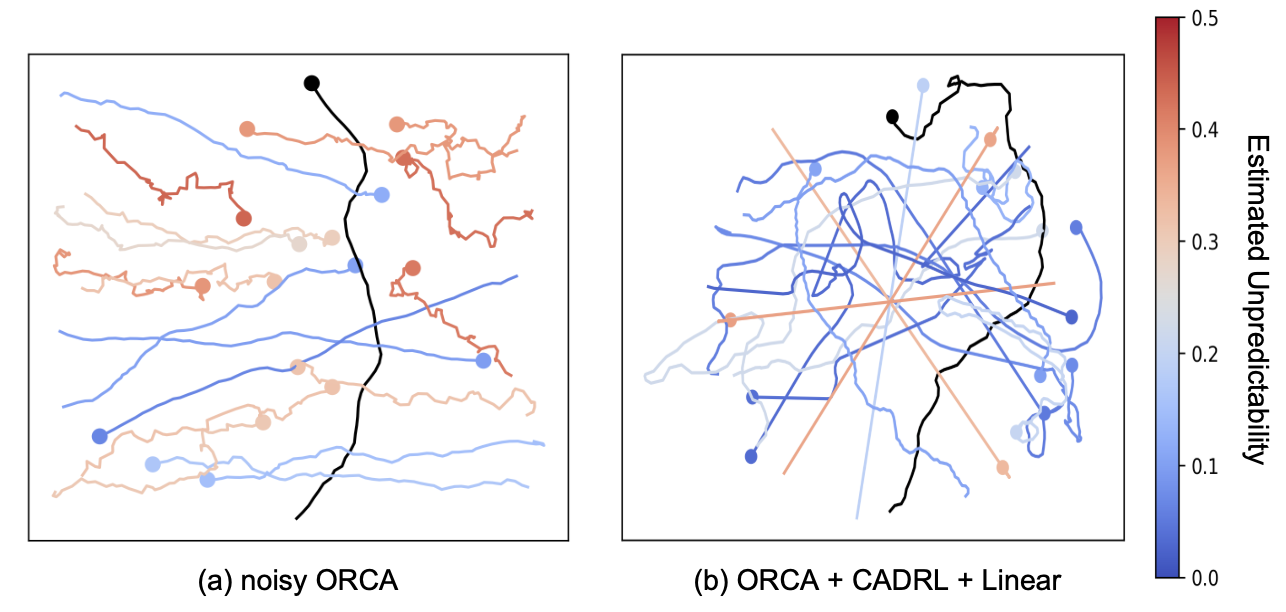}
    \caption{Estimated unpredictability values for (a) \ORCAeps{} pedestrians and (b) ORCA, CADRL, and Linear pedestrians. Lines indicate paths of the agent, and circles indicate ending positions. The robot is colored black, and the pedestrians are colored according to their average estimated unpredictability value. Notice that pedestrians who walk haphazardly (pink pedestrians in (a)) and those who walk straight through the middle without engaging in collision avoidance maneuvers (pink pedestrians in (b)) have high associated unpredictability values. Those that behave more normally (blue pedestrians in (a) and (b)) have lower associated values.\vspace{-4mm}}
    \label{fig:unpredictability}
\end{figure}

\subsection{Training Process} \label{subsec:training}

Our uncertainty-aware RL policy is trained in a modified CrowdSim environment \citep{chen_crowd-robot_2019}, where we generate arbitrarily many pedestrians with randomized initial positions and goals. By default, pedestrians choose their action at each time step based on the ORCA policy~\citep{orca} -- a navigation strategy commonly used to model human navigation behavior~\citep{mavrogiannis_core_2021}. For a pedestrian with a preferred velocity of $v_{pref}$, an ORCA action, $\vec{a}_{ORCA} \in \{\vec{v}\in\mathbb{R}^2:||\vec{v}||_2\leq v_{pref}\}$, comprises $x$ and $y$ velocities and makes progress towards a goal while avoiding collisions with other agents. To generate quantifiable deviations from this policy and systematically produce highly-heterogenous pedestrians for training, we augment the policy with Gaussian noise. Each pedestrian is instantiated with a deviation value $\rho \sim U(0, \rho_{max})$ for $\rho_{max} \in [0, 1]$, which represents how much the pedestrian deviates from the default ORCA policy. At each time step, the pedestrian takes an action 
$\vec{a} = (1-\rho)\vec{a}_{ORCA} + \rho \vec{a}_{rand}$, where $\vec{a}_{rand} \sim N(\vec{0}_2,  v_{pref} I_2)$ is a 2D Gaussian-random action. We call this noisy policy \textit{\ORCAeps{}} to differentiate it from the standard ORCA policy. The left plot in Figure \ref{fig:unpredictability} provides one example of \ORCAeps{} pedestrians. We intentionally do not ensure that this action is collision-free and rational, as real people may take actions that appear irrational and result in collision.

We found that successfully training an RL policy on \ORCAeps{} pedestrians is not trivial. These pedestrians generate spurious signals from their random motion, making it difficult for the robot to simultaneously learn how to exploit behavioral patterns in ORCA while also avoiding the unpredictable deviations from ORCA. To overcome this problem, we trained the RL policies using curriculum training, starting with standard ORCA pedestrians and gradually increasing the difficulty of the navigation scenario throughout the training process. Specifically, we increased the maximum deviation value ($\rho_{max}$) of pedestrians by 0.1 every 2,000~training episodes. We ran 12,000 episodes, concluding training with a maximum deviation value of $\rho_{max} = 0.5$. 

\subsection{Model Architecture} \label{subsec:model}

We assume the robot has access to its own position, velocity, radius, orientation, preferred velocity, and goal position. We also assume the robot has access to the position, velocity, and radius of each pedestrian that has been observed by the robot over time. We choose to use only these observations because they can be readily obtained by physical robots navigating around people in the real world. 

Our RL policy is trained using a value iteration algorithm with the value network shown in Figure~\ref{fig:nn}. Notably, we add a layer of multi-layer perceptrons (MLP$_1$), which we train separately from the rest of the network such that MLP$_1$ predicts deviation values $\hat{\rho}$ for each pedestrian while the rest of the network estimates the value function using ground truth $\rho$ values. We train these components separately so the value network learns how to utilize the quantified deviation of each pedestrian from the ORCA policy, as opposed to simply learning latent features of \ORCAeps{} pedestrians. The purpose of MLP$_1$, which we refer to as the \textit{uncertainty estimation network}, is to quantify deviation of \textit{any} observed pedestrian (not just \ORCAeps{} ones) from the ORCA policy. See Figure \ref{fig:unpredictability}b for sample unpredictability estimations for pedestrians operating under previously unseen policies.

\subsection{Reward Function} \label{subsec:reward}

The reward function used to train our RL policies encourages the robot to reach its goal while maintaining social norms and avoiding collisions with people. In an environment with $n$ pedestrians, where $d_i$ is the distance from the robot to the $i$th person and $d_g$ is the distance from the robot to its goal, the default (i.e., $\rho$-independent) reward is:
\begin{equation*}
\scalebox{0.85}{
$r = k_{succ}H(-d_g) 
+ k_{coll}\sum_{i=1}^n H(-d_i) + k_{disc}\sum_{i=1}^n \min\Bigl\{0,\ d_i-d_{disc}\Bigr\},$
}
\end{equation*}
where $H$ is the step function and $d_{disc} = 0.1$ is a constant referred to as \textit{discomfort distance}. In this function, the first term rewards the robot for reaching its goal, the second penalizes it for colliding with a person, and the third encourages it to maintain a comfortable distance from each person. 

To incorporate the intuition of avoiding close interactions with unpredictable pedestrians while freely navigating around predictable ones, we modify the discomfort distance in the reward function to be $\rho$-dependent. Given a deviation value $\rho_i$ for the $i$th pedestrian, their discomfort distance in the modified function is $d_{disc}(\rho_i) = a \rho_i + b$, where $a=1.0$ and $b=0.2$ for our experiments. Everything else from the initial reward function remains unchanged. We call this $\rho$-dependent reward function the \textit{modified reward function}.

\section{Experimental Evaluation}
\label{sec:results}

\begin{figure}
    \centering
    \captionsetup{width=\columnwidth}
    \includegraphics[width=\columnwidth]{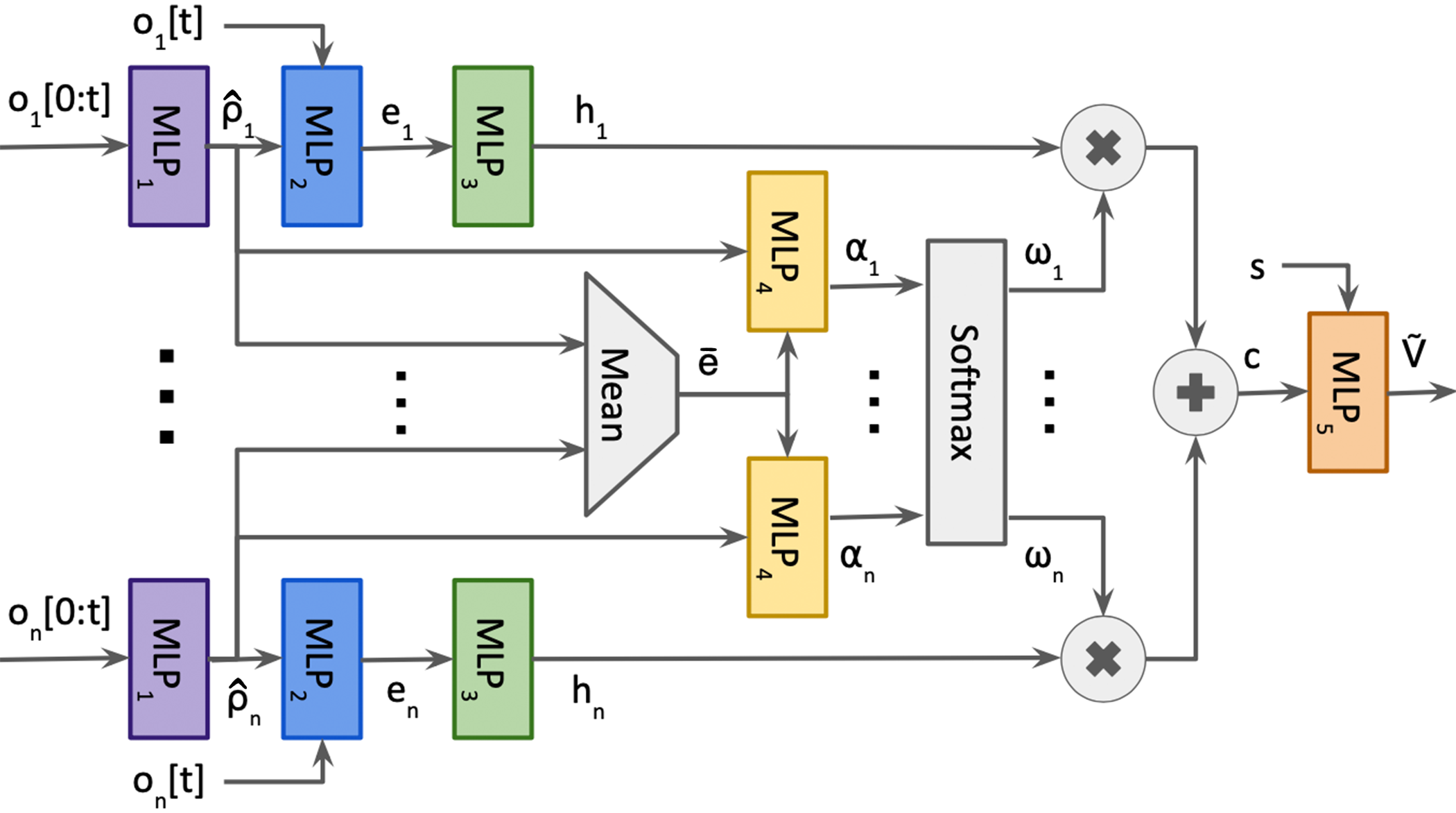}
    \caption{Our augmented value network. Given a history of observations for pedestrian $i$, $MLP_1$ estimates the ORCA policy deviation $\hat{\rho}_i$ associated with that pedestrian. This $\hat{\rho}_i$ is combined with the current observation of the pedestrian and passed into $MLP_2$. The rest of the network generates a compact representation, $c$, of the entire set of pedestrians, which is combined with the robot's state, $s$, to obtain an estimate of the value function, $\tilde{V}(s, o_1[0:t],\hdots,o_n[0:t])$.
    \vspace{-4mm}}
    \label{fig:nn}
\end{figure}

We conduct two simulated ablation studies of our RL policy to analyze how the policy behaves in various situations. We also implement our RL policy on a physical robot and discuss some takeaways from our hardware experiment.

\subsection{Simulation Experimental Setups}
\label{subsec:experiments}

\begin{figure*}[h!]
\includegraphics[width=0.7\columnwidth]{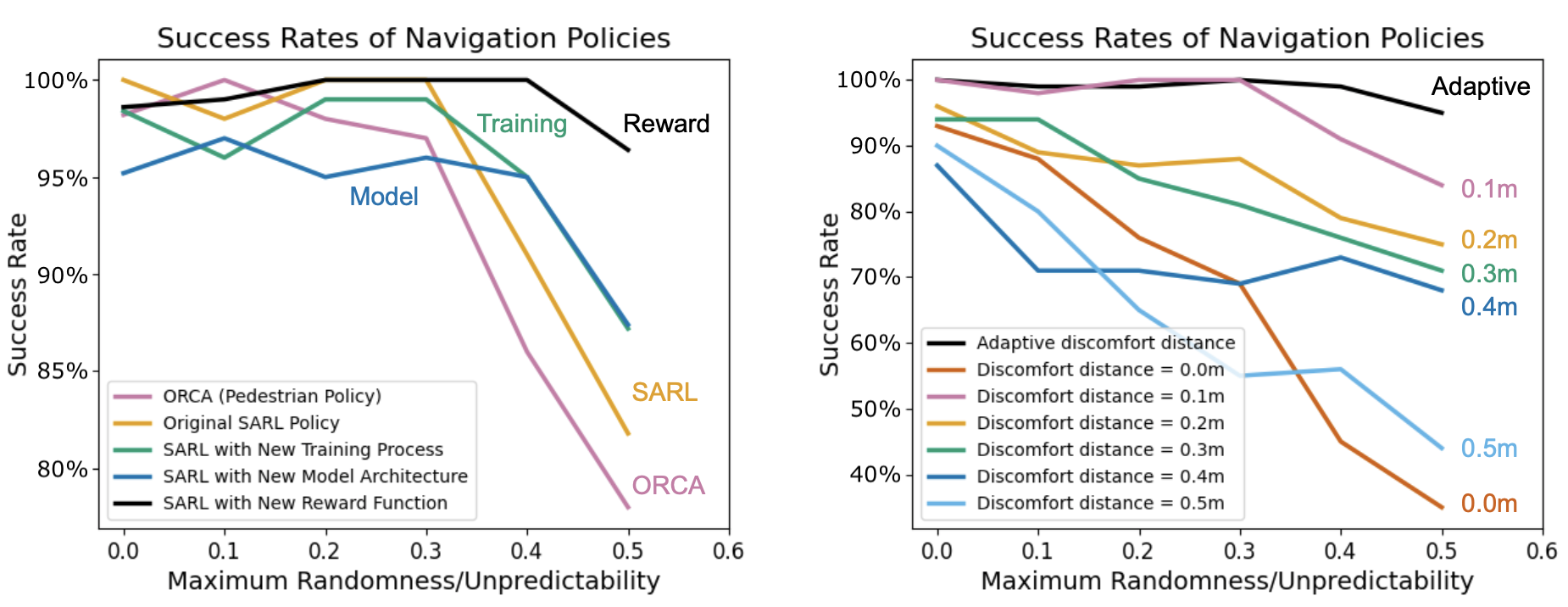}
\vspace{-3mm}
\centering
\caption{Success rates of various policies navigating among \ORCAeps{} pedestrians as pedestrian unpredictability increases across 500 trials. (Left) Ablation study of our uncertainty-aware policy. Performance improves as uncertainty is integrated by successively modifying the \textit{Training} process, the \textit{Model} architecture, and the \textit{Reward} function. (Right) Comparison of our uncertainty-aware policy against standard SARL policies with a variety of fixed discomfort distance parameters.
\vspace{-3mm}}
\label{fig:ablation_and_disc_dists}
\end{figure*} 

We evaluate our RL policies in a modified CrowdSim environment by conducting randomized episodes across six distinct categories of robot-pedestrian interactions (circle and perpendicular crossing, oncoming and outgoing flow, single and perpetual random goals) comprising a superset of scenarios presented in prior works~\cite{gao_evaluation_2022, 9802981}. We perform two sets of experiments in this environment using (i)~\ORCAeps{} pedestrians and (ii)~pedestrians operating according to more realistic but varied policies. \ORCAeps{} pedestrians are quantifiably diverse, allowing us to evaluate our policies under a formalized concept of domain shift. However, because real people do not move with Gaussian random noise, \ORCAeps{} pedestrians cannot be expected to reflect true deviations in human behavior. For more realistic deviations, we design experiments where pedestrians operate under standard ORCA \cite{orca}, CADRL \cite{cadrl}, and Linear policies with a wide range of different parameters. In these experiments, each pedestrian is uniformly randomly assigned one of these three policies, and the parameters for their assigned policy are uniformly randomized as well. These pedestrians present realistic but previously unseen and highly heterogenous pedestrian behavior in our suite of scenarios.

\subsection{Performance Metrics}
\label{subsec:metrics}

To compare navigation policies, we evaluate the robot's ability to efficiently navigate to its goal while preserving the safety and comfort of surrounding pedestrians using the following metrics: (1)~\textit{Success rate}: percentage of trials where the robot successfully reaches its goal within 30~seconds. (2)~\textit{Timeout rate}: percentage of trials where the robot fails to reach the goal in the allotted time. (3)~\textit{Collision rate}: percentage of trials where the robot collides with at least one pedestrian. (4)~\textit{Relative navigation time}: time required to navigate to the goal (relative to the fastest time). (5)~\textit{Relative path length}: distance traveled by the robot to its goal (relative to the shortest path). (6)~\textit{Number of collisions}: total number of collisions between the robot and any pedestrian across all trials. (7)~\textit{Personal space cost}: overall personal space cost incurred by the robot (as defined by \cite{rachel_kirby} with parameters from \cite{liu_simulating_2019}). (8)~\textit{Personal space violation}: percentage of time spent within the personal space of a pedestrian (as defined by \cite{beermann_connection_2023}). (9)~\textit{Intimate space violation}: percentage of time spent within the intimate space of a pedestrian (as defined by \cite{beermann_connection_2023}). 
Metrics 1~--~5 describe robot path efficiency, while metrics 6~--~9 quantify the comfort of nearby pedestrians.

\begin{table*}[t]
\centering
\vspace{2mm}
\scalebox{0.75}{
\begin{tabular}{|l||ccccc|cccc|}
\hline
 Navigation     & Success       & Timeout       & Collision     &  Relative             & Relative              &  Number of    & Personal              & Personal Space            & Intimate Space            \\ 
 Policy         & Rate          & Rate          & Rate          &  Navigation Time      & Path Length           &  Collisions   & Space Cost            & Violation                 & Violation                 \\
\hline
 ORCA           &     69\%      &   \B{0\%}     &      31\%     & \B{1.09}              &  \B{1.04}             &    156        & 24.5                  & 31.4\%                    & 12.4\%                    \\
                &               &               &               & (\B{1.32} / \B{1.56}) & (\B{1.24} / \B{1.36}) &               & (91.9 / 124.7)        & (66.2\% / 73.2\%)         & (37.3\% / 44.3\%)         \\
 \hline
 \hline
 SARL           &     74\%      &  \B{0\%}      &      26\%     &    1.40               &   \B{1.27}            &    153        & 10.4                  & 17.0\%                    & 3.7\%                     \\
                &               &               &               & (1.92 / 2.06)         & (1.64 / 1.72)         &               & (47.2 / 58.2)         & (47.6\% / 55.2\%)         & (18.5\% / 23.5\%)         \\
 \hline 
 Training       &     85\%      &     2\%       &      13\%     &    1.41               &   1.30                &    39         & 13.5                  & 13.5\%                    & 2.4\%                     \\
                &               &               &               & (2.11 / 2.22)         & (1.79 / 1.92)         &               & (49.5 / 57.6)         & (38.5\% / 42.7\%)         & (11.9\% / 16.0\%)         \\
 \hline
 Model      &     75\%      &     5\%       &      20\%     &    1.42               &   1.30                &    84         & 13.4                  & 13.1\%                    & 2.9\%                     \\
                &               &               &               & (2.04 / 2.27)         & (1.70 / 1.81)         &               & (61.0 / 76.8)         & (38.2\% / 42.9\%)         & (13.7\% / 15.0\%)         \\
\hline
 Reward     &   \B{87\%}    &     2\%       &    \B{11\%}   &  \B{1.39}             &   1.31                &  \B{28}       & \B{7.7}               & \B{10.8\%}                & \B{1.7\%}                 \\
                &               &               &               & (\B{1.77} / \B{1.86}) & (\B{1.61} / \B{1.66}) &               & (\B{32.7} / \B{41.3}) & (\B{31.3\%} / \B{36.1\%}) & (\B{9.3\%} / \B{12.3\%})  \\
\hline
\end{tabular}
}
\vspace{-2mm}
\caption{Ablation study of our uncertainty-aware social navigation policy and two baseline policies (ORCA and SARL) on 100 randomized scenarios with up to 20 pedestrians running ORCA, CADRL, and Linear policies. In addition to average values, we report (10\% CVaR / 5\% CVaR) values to describe expected performance on the hardest 10\% and 5\% of all trials. We see that integrating uncertainty-awareness allows the policy to generalize better to novel and challenging situations.\vspace{-4mm}
}
\label{table:diff_policies-long}
\end{table*}

\subsection{Simulation Results \& Analysis}

Our ablation studies analyze the cumulative impact of modifying the training process~(\S \ref{subsec:training}), the model architecture~(\S \ref{subsec:model}), and the reward function~(\S \ref{subsec:reward}) of the original SARL policy. Since these modifications have sequential dependencies, we define our ablation study as follows: the original policy with no modification is referred to as \textit{SARL}, the policy with only the modified training process is referred to as \textit{Training}, the policy with both the modified training process and model architecture is referred to as \textit{Model}, and the policy with all three modifications (training process, model architecture, and reward function) is referred to as \textit{Reward} or ``our full uncertainty-aware policy.'' We show that the combination of these three policy modifications improves policy performance on particularly complex scenarios. 

\subsubsection{Ablation Study on \ORCAeps{} Pedestrians} \label{sec:ablation}

We run 500 trials with increasingly noisy pedestrians to quantify the policies' performance under domain shift. In the left plot of Figure~\ref{fig:ablation_and_disc_dists}, we see that ORCA and all variations of the socially-aware RL policy perform comparably when the pedestrians navigate with only small deviations from ORCA (i.e., $\rho_{max} < 0.3$). However, as the maximum randomness of the pedestrians increases, the performance of all of the policies drops significantly except for that of \textit{Reward}. We see the most significant performance drops for ORCA and standard SARL, which is expected because they implicitly expect pedestrians to behave according to the original ORCA policy. We also see a reasonably large drop for \textit{Training} and \textit{Model}, which indicates that the modified reward function in \textit{Reward} is crucial for learning robot responses that generalize well to significant changes in pedestrian behavior.

\subsubsection{Comparing Different Discomfort Distances} \label{sec:discomfort}

Since the greatest performance improvement comes from using a $\rho$-dependent discomfort distance in the reward function, it is natural to suspect that we can improve performance by simply tuning the constant $\rho$-\textit{independent} distance in the original reward function. We explore this in the right plot of Figure \ref{fig:ablation_and_disc_dists}. The policy trained with a discomfort distance of 0m frequently collides with pedestrians, while policies trained with discomfort distances greater than 0.1m often time out from being too cautious, resulting in low success rates. The policy trained with a discomfort distance of 0.1m does well when $\rho_{max} \leq 0.3$ but does not generalize to higher variability. Our adaptive discomfort distance results in the best performance across all levels of pedestrian unpredictability. 

\subsubsection{Ablation Study on Diverse, Realistic Pedestrians} \label{sec:diverse}


We further evaluate the impact of uncertainty integration on navigation by conducting trials with more realistic pedestrian behavior and summarize our results in Table~\ref{table:diff_policies-long}. 
These trials contain a mix of up to twenty pedestrians in 100 randomized scenarios. Each pedestrian is uniformly randomly assigned to navigate according to the \textit{standard ORCA}, \textit{CADRL}, or \textit{Linear} policy. In addition to average performance, we report Conditional Value at Risk (CVaR),  reflecting the average performance on the hardest $10\%$ and $5\%$ of all trials, to quantify performance in particularly challenging situations.

\textbf{ORCA} consistently has lower success rates, higher collision rates, shorter navigation times, and shorter path lengths compared to the RL policies (i.e.,~SARL and its modifications). 
Upon visually inspecting trials, we see that the ORCA policy takes very direct and aggressive paths towards the goal, making little effort to maintain pedestrian comfort. This is reflected in the high personal space cost, personal space violation, and intimate space violation. Thus, ORCA is unsuitable to use as a \textit{social} navigation policy baseline, so we focus the rest of our analysis comparing our policy modifications to the SARL baseline instead.

\textbf{SARL} performs significantly worse in the scenarios presented in Table~\ref{table:diff_policies-long} than in those containing \ORCAeps{} pedestrians presented in Figure~\ref{fig:ablation_and_disc_dists}. This underscores the difficulty for learning-based algorithms to overcome domain shift, even in simulated environments. 
The standard SARL policy exhibits lower success rates, higher collision rates, far more total collisions, and more personal and intimate space violations compared to all modified policies (\textit{Training}, \textit{Model}, and \textit{Reward}). 
The performance of each of these three modified policies also vary in turn. 

\textbf{Training} reduces the number of collisions by 75\% relative to standard SARL. This indicates that including noisy and potentially unrealistic behavior in the training distribution (i.e.,~performing data augmentation) can improve performance in more realistic scenarios. However, \textit{Training} has slightly worse path efficiency performance compared to the original policy. By visually inspecting some of the trials, we infer that this policy learns responses that are overly cautious toward all pedestrians, including those that behave predictably. This indicates that simply training policies with unpredictable pedestrians is not sufficient for learning efficient yet generalizable behavior. 


\textbf{Model}, despite having access to additional uncertainty information, generally performs worse than \textit{Training}. In particular, \textit{Model} collides twice as often as \textit{Training}. This may indicate that \textit{Model} overfits to spurious patterns involving the unpredictability feature $\rho$ of \ORCAeps{} pedestrians, and thus fails to learn information about unpredictability that generalizes to more realistic human policies.

\textbf{Reward} exhibits the best performance overall. Unlike \textit{Training}, \textit{Reward} is moderately more efficient than standard SARL in hard cases. \textit{Reward} also recovers from the high collision rate exhibited by \textit{Model}, indicating that the modified reward function is critical in learning generalizable behavior from the unpredictability feature $\rho$. Overall, \textit{Reward} reduces the number of collisions by 82\% relative to standard SARL and reduces the time spent in pedestrians' personal space by 6.2~percentage points on average and by 16.3 to 19.1~percentage points for hard cases. It also reduces time spent in their intimate space by 2.0~percentage points on average and by 9.2 to 11.2~percentage points for hard cases. 

Thus, our proposed modifications are able to generate RL policies that notably improve pedestrian safety and comfort in particularly challenging scenarios while still maintaining good expected performance overall.

\subsection{Robotic Experiment}

To evaluate the ability of our uncertainty-aware policy to identify and respond to novel situations in the real world, we implement our full uncertainty-aware policy on a TurtleBot with a single, front-facing RGBD camera. We use a pre-trained YOLOv3 neural network \citep{yolo} to detect pedestrians and design an algorithm to obtain their state information and track them over time. We compare the behavior of the standard SARL policy to our full uncertainty-aware policy in three familiar scenarios containing predictable pedestrians (Crossing, Passing, and Overtaking) and two unfamiliar scenarios containing unpredictable pedestrians (Standing and Stopping). Since ORCA pedestrians seen during training are constantly moving, Standing and Stopping behaviors are not part of the training distribution, so policies trained on ORCA pedestrians are limited in their ability to accurately predict future actions of pedestrians exhibiting these behaviors. Hardware evaluations on more unusual and haphazard pedestrian motions (e.g., zig-zagging) are not performed because the robot’s responsiveness is limited by long image-processing times and a narrow camera field-of-view. A more capable hardware platform would support more extensive experimentation in future work. Regardless, we find that our full uncertainty-aware policy behaves similarly to SARL in familiar situations, while our policy successfully identifies unpredictable pedestrians and keeps a larger distance from them in novel situations, as compared to the original SARL policy. Therefore, the key high-level behaviors of our approach transfer from simulation to hardware in these scenarios. Video clips for the Crossing and Stopping episodes of our robotic experiment are included in our video\footnote{A video summarizing our methods and results is available on YouTube: https://youtu.be/9IDhXvCC58w.}.
\section{Extensions to other RL policies}
\label{sec:extensions}

In our work, we demonstrate the need to train and evaluate RL-based social navigation policies with the consideration of domain shift. While we focus on modifying SARL~\cite{chen_crowd-robot_2019}, our proposed modifications are not restricted to this particular policy. Even given significant advances in model architectures for RL-based social navigation, the overall framework of many policies remains conducive to these modifications.

\textbf{Training process:} We develop a curriculum training process, where the RL policy is initially trained as normal. As training progresses, Gaussian noise is increasingly added to the pedestrians' actions. While there are many approaches for modeling pedestrian behavior during RL policy training, the training process of any policy can be modified in this way as long as the action values for pedestrians in the environment are accessible. Regardless of how pedestrian actions are determined, each pedestrian can be initialized with a corresponding deviation value $\rho$ and their actions adjusted with $\rho$-dependent Gaussian random noise.

\textbf{Model architecture:} We propose a modification to the RL policy model architecture by (1) training an uncertainty estimation network and (2) incorporating the uncertainty estimations as agent-level features in the observation space. The uncertainty estimation network can be developed entirely independently from the RL policy, so this component is completely policy-agnostic. For the uncertainty estimations to be seamlessly incorporated into the original RL policy, the observation space of the original policy must contain agent-level features (e.g., position and velocity values for each nearby pedestrian). This is true for many existing policies. Additional modifications would have to be made for end-to-end RL policies that operate directly on raw sensor measurements (e.g., images or 2D lidar).

\textbf{Reward function:} We propose a modification to the reward function that encourages the robot to maintain additional space around pedestrians that deviate from the assumed model of pedestrian behavior. To make this same modification in other RL policies, their reward function simply must contain some notion of ``safety space,'' ``discomfort distance,'' or ``clearance'' that captures a sense of maintaining proper distance from pedestrians. If this is the case, the modification of increasing each agent's discomfort distance based on their deviation value $\rho$ is subsequently straightforward, though specific constants may need to be tuned for the particular model of interest.

\section{Conclusions}
\label{sec:conclusion}

In this work, we articulate the domain shift problem for RL policies in social robot navigation and present an approach that improves generalizability of RL policies to novel scenarios while maintaining their efficiency in familiar ones. We find that SARL~\citep{chen_crowd-robot_2019} generalizes poorly to significant deviations in pedestrian behavior, thereby presenting serious concerns for pedestrian safety and comfort in a real-world deployment. We posit that for socially-aware RL policies to be viable in real-world mobile robots, these policies must recognize when people deviate from the (implicitly) assumed pedestrian model and take appropriate caution. We present effective methods for modifying the training process, the model architecture, and the reward function of SARL that substantially improve the generalizability of the policy. Comparing our modified policy to the original SARL policy on randomized scenarios containing realistic ORCA and non-ORCA human policies, our modifications reduce the number of collisions by 82\% and reduce the proportion of time spent in the pedestrians' personal space by 16~percentage points for the hardest 10\% of all trials and by 19~percentage points for the hardest 5\% of trials. This increase in pedestrian comfort is achieved while maintaining similar navigation times and path lengths. We also discuss how these same modifications can be applied to other socially-aware RL policies.

While we believe our work takes an important step toward enabling the deployment of socially-aware RL policies on mobile robots, there are some limitations that should be addressed. First, we modify the reward function to encourage the robot to maintain \textit{greater space} between itself and unpredictable pedestrians. While this heuristic for caution is reasonable in many situations, it is less effective in tight spaces, where the robot is unable to maintain such a distance. It would be interesting to explore other heuristics, such as slowing down, speeding up, or some combination of adjusting distance and speed when approaching unpredictable pedestrians. Another limitation is that we use the ORCA policy as our primary pedestrian model for this study because this is the model commonly used in other RL-based social navigation work. However, this model is relatively simplistic. It would be interesting to train RL policies using other models of pedestrian behavior and evaluate generalizability to even more diverse and realistic pedestrian scenarios. 

\addtolength{\textheight}{-4cm}   









\bibliographystyle{IEEEtran}
\bibliography{IEEEabrv, references}


\newpage
\newpage
\section{Appendix}




\subsection{Additional RL Policy Details}

\subsubsection{Observation Space}

The observation space of the RL policy consists of the full state of the robot and the observable state of each of the $n$ pedestrians. The full state of the robot is given by
\[
\bm{o}^{(r)} = 
\begin{bmatrix}
p_x^{(r)} & p_y^{(r)} & v_x^{(r)} & v_y^{(r)} & r^{(r)} & g_x^{(r)} & g_y^{(r)} & v_{pref}^{(r)} & \theta^{(r)}
\end{bmatrix}
\]
and the observable state of the $i$th pedestrian is given by
\[
\bm{o}^{(h_i)} = 
\begin{bmatrix}
p_x^{(h_i)} & p_y^{(h_i)} & v_x^{(h_i)} & v_y^{(h_i)} & r^{(h_i)}
\end{bmatrix},
\]
where $(p_x,p_y)$ is the current position, $(v_x,v_y)$ is the velocity, $r$ is the radius, $(g_x,g_y)$ is the goal position, $v_{pref}$ is the preferred velocity, and $\theta$ is the turning angle corresponding to the robot ($r$) or $i$th human ($h_i$). The current position, velocity, goal position, and turning angle are all measured with respect to a fixed world frame. The radius is measured assuming that the robot and each pedestrian can be represented using a circle with a fixed radius. The preferred velocity is the maximum possible speed of each agent, which is the speed it would travel at if nothing is obstructing its path. 


\subsubsection{Action Space}

The robot receives linear and rotational velocity commands as actions under the assumption that the velocity of the robot can be achieved instantly after the action command is received. The action space is discretized into $16$ rotations evenly spaced in the range $[0,2\pi)$ and $5$ speeds exponentially spaced in the range $(0,v_{pref}^{(r)}]$ such that the full set of speeds of the robot is
\[
\left\{\left(\frac{e^{i/5} - 1}{e - 1}\right)v_{pref}^{(r)},\ i=[1,5]\right\}.
\]

The robot is also able to receive a stop command, resulting in $81$ possible discrete actions. 

\subsubsection{Value Network Architecture} \label{appendix:model}

The parameters used in our model architecture are given in Table \ref{table:nn}. For the uncertainty estimation network (MLP$_1$), we conducted a parametric depth study (from 0 to 18 hidden layers) to see how the size of the network improved inference performance. Regarding the number of hidden layers, the models' performance improved as the number of hidden layers increased from 0 to 5, but degenerated with more than 9 hidden layers.

\subsubsection{Network Input Preprocessing} \label{appendix:features}

We preprocess the pedestrian observations ($o_1$ through $o_n$ in Figure \ref{fig:nn}) before using them as input to the value network. For the uncertainty estimation network (MLP$_1$), we explored a number of different possible input features. In particular, we considered pedestrian position, velocity, acceleration, distance to robot, speed, magnitude of acceleration, curvature of  trajectory, distance to nearest neighbors, and number of pedestrians within a neighborhood of the pedestrian across a number of different time steps. We decide to use only scalar speed and acceleration as our input features, ranging from 1 to 20 time steps, where each time step is 0.25 seconds.

To preprocess the input to MLP$_2$, we first express the observation of the robot at a given time step as
\[
\bm{O} =
\begin{bmatrix}
\bm{o}^{(r)} & \bm{o}^{(h_1)} \\
\vdots & \vdots \\
\bm{o}^{(r)} & \bm{o}^{(h_n)} \\
\end{bmatrix}.
\]

Once the observation is expressed in this form, it is useful to rotate it to a robot-centric coordinate frame. Consider the $i$th row of the matrix $\bm{O}$. The rotated version of this observation is given by the following vector:

\small
\[
\Tilde{\bm{o}}_i =
\Bigl[ d_g \ \ \Tilde{v}_x^{(r)} \ \ \Tilde{v}_y^{(r)} \ \ r^{(r)} \ \ v_{pref}^{(r)} \ \ \theta^{(r)} \ \ \Tilde{p}_x^{(h_i)} \ \ \Tilde{p}_y^{(h_i)} \ \ \Tilde{v}_x^{(h_i)} \ \ \Tilde{v}_y^{(h_i)} \ \ r^{(h_i)} \ \ d_{h_i} \ \ r_{tot} \Bigr].
\]
\normalsize

The variables in the rotated vector are provided in Table \ref{table:observation}. These rotated vectors are used as input to MLP$_2$ in our value network model architecture (Figure \ref{fig:nn}).

\subsubsection{Reward Function} \label{appendix:reward}

The parameters used in our reward function are given in Table \ref{table:reward}.

\begin{table}[h!]
\centering
\begin{tabular}[width=\columnwidth]{|c | c|} 
\hline
Parameter & Value \\ [0.5ex]
\hline \hline
Success Reward ($k_{succ}$) & $1.0$ \\ 
\hline
Collision Penalty ($k_{coll}$) & $-0.25$ \\
\hline
\hspace{5mm} Discomfort Penalty Factor ($k_{disc}$) \hspace{5mm} & $0.125$ \\
\hline
Discomfort Distance Slope ($\alpha$) & $1.0$ \\
\hline
Discomfort Distance Intercept ($\beta$) & $0.2$ \\
\hline
\end{tabular}
\caption{RL policy reward function parameters.}
\label{table:reward}
\end{table}

\begin{table*}[h!]
    \centering
    \begin{tabular}[width=\columnwidth]{|c | c|} 
    \hline
    MLP & Layers \\ [0.5ex] 
    \hline \hline
    1 & \scriptsize Lin(t, 150) $\rightarrow$ \scriptsize ReLU $\rightarrow$ \scriptsize Lin(150, 100) $\rightarrow$ \scriptsize ReLU $\rightarrow$ \scriptsize Lin(100, 100) $\rightarrow$ \scriptsize ReLU $\rightarrow$ \scriptsize Lin(100, 100) $\rightarrow$ \scriptsize ReLU $\rightarrow$ \scriptsize Lin(100, 50) $\rightarrow$ \scriptsize ReLU $\rightarrow$ \scriptsize Lin(50, 1) \\ 
    \hline
    2 & Linear(14, 150) $\rightarrow$ ReLU $\rightarrow$ Linear(150, 100) $\rightarrow$ ReLU \\ 
    \hline
    3 & Linear(100, 100) $\rightarrow$ ReLU $\rightarrow$ Linear(100, 50) \\
    \hline
    4 & Linear(200, 100) $\rightarrow$ ReLU $\rightarrow$ Linear(100, 100) $\rightarrow$ ReLU $\rightarrow$ Linear(100, 1) \\
    \hline
    5 & \small Lin(56, 150) $\rightarrow$ \small ReLU $\rightarrow$ \small Lin(150, 100) $\rightarrow$ \small ReLU $\rightarrow$ \small Lin(100, 100) $\rightarrow$ \small ReLU $\rightarrow$ \small Lin(100, 1) \\
    \hline
    \end{tabular}   
    \captionsetup{width=\linewidth}
    \caption[Neural network layer parameters]{The five MLPs in the value network are composed of Rectified Linear Unit (ReLU) activation functions and linear layers, whose input and output sizes are as shown.}
    \label{table:nn}
\end{table*}

\begin{table*}[h!]
    \centering
    \begin{tabular}[width=\columnwidth]{|c|c|c|} 
    \hline
    Variable & Description & Definition \\ [0.5ex] 
    \hline \hline
    $d_{gx}$ & Horizontal distance of robot to goal & $g_x^{(r)} - p_x^{(r)}$ \\
    $d_{gy}$ & Vertical distance of robot to goal & $g_y^{(r)} - p_y^{(r)}$ \\
    $d_g$ & Total distance of robot to goal & $\sqrt{d_{gx}^2 + d_{gy}^2}$ \\
    $\phi$ & Angle between robot and goal position & $\tan^{-1}\left(\frac{d_{gy}}{d_{gx}}\right)$ \\
    $\Tilde{v}_x^{(r)}$ & Transformed horizontal velocity of robot & $v_x^{(r)}\cos(\phi) + v_y^{(r)}\sin(\phi)$ \\
    $\Tilde{v}_y^{(r)}$ & Transformed vertical velocity of robot & $v_y^{(r)}\cos(\phi) - v_x^{(r)}\sin(\phi)$ \\
    $\Tilde{p}_x^{(h_i)}$ & Transformed horizontal position of pedestrian & $\Bigl(p_x^{(h_i)} - p_x^{(r)}\Bigr)\cos(\phi) + \Bigl(p_y^{(h_i)} - p_y^{(r)}\Bigr)\sin(\phi)$ \\
    $\Tilde{p}_y^{(h_i)}$ & Transformed vertical position of pedestrian & $\Bigl(p_y^{(h_i)} - p_y^{(r)}\Bigr)\cos(\phi) - \Bigl(p_x^{(h_i)} - p_x^{(r)}\Bigr)\sin(\phi)$ \\
    $\Tilde{v}_x^{(h_i)}$ & Transformed horizontal velocity of pedestrian & $v_x^{(h_i)}\cos(\phi) + v_y^{(h_i)}\sin(\phi)$ \\
    $\Tilde{v}_y^{(h_i)}$ & Transformed vertical velocity of pedestrian & $v_y^{(h_i)}\cos(\phi) - v_x^{(h_i)}\sin(\phi)$ \\
    $d_{h_i}$ & Total distance of robot to pedestrian & $\sqrt{\Bigl(p_x^{(h_i)} - p_x^{(r)}\Bigr)^2 + \Bigl(p_y^{(h_i)} - p_y^{(r)}\Bigr)^2}$ \\
    $r_{tot}$ & Combined radius of robot and pedestrian & $ r^{(r)} + r^{(h_i)}$ \\
    \hline
    \end{tabular}   
    \captionsetup{width=\linewidth}
    \caption{The variables in the robot-centric observation for pedestrian $i$ used as input to MLP$_2$ in the value network.}
    \label{table:observation}
\end{table*}

\subsubsection{Training Details} \label{appendix:training}

Our RL policy was trained using the value network algorithm described in Algorithm \ref{alg:value} with the mean squared error (MSE) loss function. Table \ref{table:train} summarizes the parameters chosen for training. Note that the value of epsilon used in the epsilon-greedy policy is given by
\[
\epsilon =
\begin{cases}
\epsilon_{start} + \left(\frac{\epsilon_{end} - \epsilon_{start}}{\tau_{\epsilon}}\right)\text{episode} &\text{if } \text{episode} \leq \tau_{\epsilon} \\
\epsilon_{end} &\text{otherwise}
\end{cases}.
\]

\begin{table}[h!]
\centering
\begin{tabular}[width=\columnwidth]{|c | c|} 
\hline
Hyperparameter & Value \\ [0.5ex] 
\hline \hline
Discount Factor ($\gamma$) & $0.9$  \\ 
\hline
Learning Rate ($\alpha$) & $0.001$  \\ 
\hline
Batch Size ($M$) & $100$  \\ 
\hline
Number of Training Episodes ($N$) & $12000$  \\ 
\hline
Target Update Interval ($\tau_{TU}$) & $50$  \\ 
\hline
Initial Epsilon ($\epsilon_{start}$) & $0.5$  \\ 
\hline
Final Epsilon ($\epsilon_{end}$) & $0.1$  \\ 
\hline
Epsilon Decay ($\tau_{\epsilon}$) & $4000$  \\ 
\hline
SGD Momentum ($\beta$) & $0.9$  \\ 
\hline
\end{tabular}
\caption{Reinforcement learning training parameters.}
\label{table:train}
\end{table}

\begin{algorithm}[t]
\caption{Value Network Algorithm}
\label{alg:value}
\begin{algorithmic}[1]
\STATE Initialize the parameters $\phi$ and $\phi'$ of the value network and target network.
\STATE Initialize the replay buffer using state transitions from the ORCA policy.
\FOR{episode $=1\rightarrow N$}
    \STATE Set $t \leftarrow 0$
    \REPEAT
        \STATE Determine the optimal action for the current state using the epsilon greedy policy:
        \[
        a_t^* \sim \pi_\theta(a_t|s_t)
        \]
        \STATE Take the optimal action, observe the state transition, and add it to the replay buffer.
        \STATE Randomly sample a batch of $M$ state transitions from the replay buffer.
        \STATE Compute the target value for each of the $M$ samples in the batch:
        \[
        y_i = r(s_i, a_i) + \gamma \hat{V}_{\phi'}^\pi(s_i'),\ i=1,\hdots,M
        \]
        \STATE Update the value function network parameters using stochastic gradient descent:
        \[
        \phi\leftarrow \phi - \alpha\frac{1}{M}\sum_{i=1}^M \frac{d}{d\phi}\ell\Bigl(\hat{V}_\phi(s_i), y_i\Bigr)
        \]
        \STATE Increment the current time step: $t\leftarrow t+1$.
    \UNTIL{$s_t$ reaches a terminal state or $t\geq t_{max}$}
    \IF{$\text{episode} \bmod \tau_{TU}$ = 0}
        \STATE Update the target network parameters: $\phi'\leftarrow\phi$.
    \ENDIF
\ENDFOR
\RETURN $\hat{V}_\phi$
\end{algorithmic}
\end{algorithm}

\subsubsection{Optimal Policy}

Given the current state $s$, the expected next state $s'$ is determined for each action $a$ in the action space, $\mathcal{A}$, by using a simple model to approximate the motion of the robot and pedestrians. The next state is then used as input to the value network to determine the value $\tilde{V}(s')$. This value is then multiplied by the discount factor, $\gamma$, and added to the reward, $r(s,a)$, for the current state and action. The action with the greatest corresponding value of $r(s,a) + \gamma \tilde{V}(s')$ is the optimal action:
\[
a^* = \argmax_{a\in\mathcal{A}}\ \Bigl(r(s,a) + \gamma \tilde{V}(s')\Bigr).
\]

Recall that the action space, $\mathcal{A}$, is discretized into $5$ speeds and $16$ rotations, along with a stop command, resulting in $81$ possible actions. Because the action space is discrete, the value of $r(s,a) + \gamma \tilde{V}(s')$ can be explicitly computed for each of these 81 actions. The discrete action corresponding to the maximum value of $r(s,a) + \gamma \tilde{V}(s')$ is then used to control the velocity of the robot.



\subsection{Additional Simulation Experiment Details}

\begin{figure}[h!]
    \centering
    \captionsetup{width=\linewidth}
    \includegraphics[width=\textwidth]{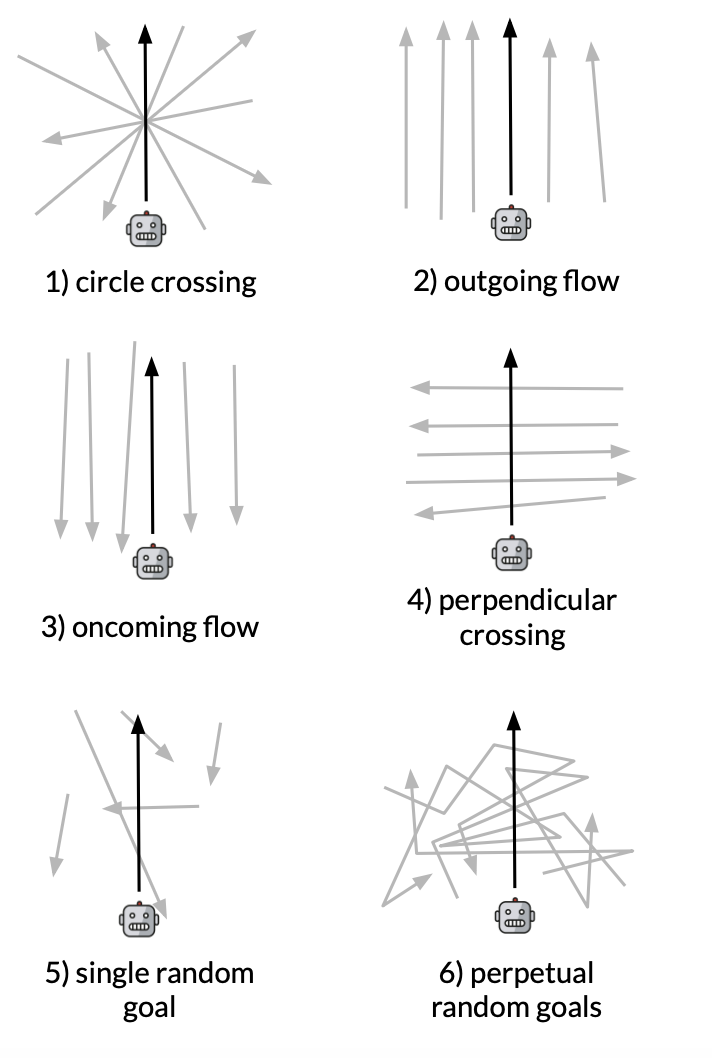}
    \caption{Example of each category of scenarios used in simulation experiments. Arrows indicate state and goal positions for the robot (black) and pedestrians (gray).}
    \label{fig:scenarios}
\end{figure}

In our CrowdSim environment, we simulate several variations of different human crossing situations, which were chosen to capture all of the primitive interactions between a robot and surrounding people \citep{gao_evaluation_2022}, \citep{9802981}. We designed the following scenarios: (1) \textit{Circle crossing}: the robot and pedestrians each start at some point on a circle and move to the opposite point. (2) \textit{Outgoing flow}: the robot and pedestrians start on the same side of a room and move to the opposite side. (3) \textit{Oncoming flow}: the robot and pedestrians start on opposite sides of a room and pass each other to reach the other side. (4) \textit{Perpendicular crossing}: the robot starts at the bottom of the room and the pedestrians start on the left and right sides. Everyone moves to the side opposite of their starting position. (5) \textit{Single random goal}: the robot moves across a room, while each pedestrian is given a random start and goal position. (6) \textit{Perpetual random goals}: each pedestrian is now given a new random goal after reaching its current goal. The scenarios are visualized in Figure \ref{fig:scenarios}.


\subsection{Additional Robotic Experiment Details} \label{appendix:hardware}

\subsubsection{Robot Platform \& Interface}

\begin{figure}[h!]
    \centering
    \captionsetup{width=\linewidth}
    \includegraphics[width=\textwidth]{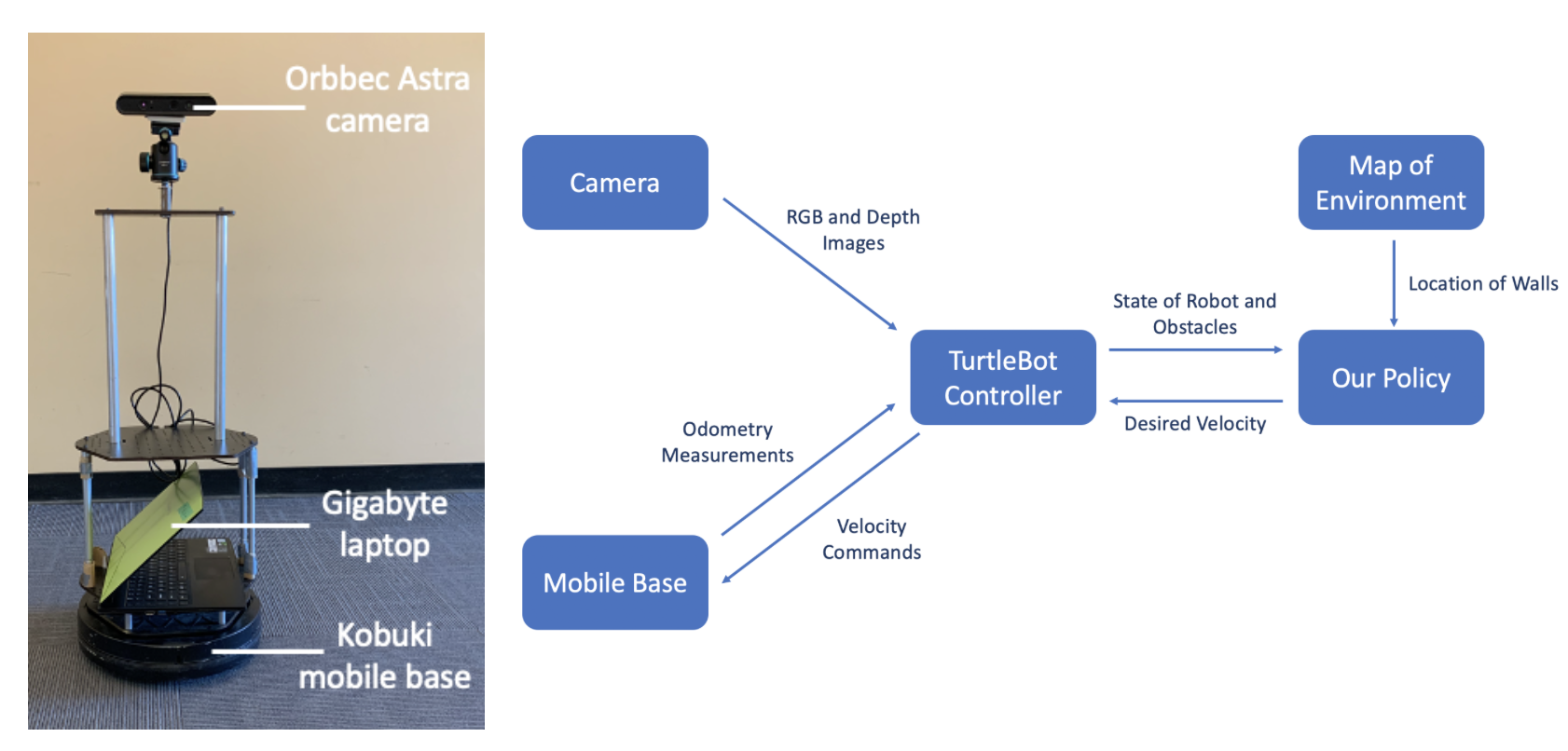}
    \caption{(Left) We used a TurtleBot as the robot platform, which is composed of an Orbecc Astra camera, a Gigabyte laptop computer, and a Kobuki mobile base. (Right) The TurtleBot controller communicates with the camera, mobile base, and our socially-aware policy to control the robot in real-world environments around pedestrians. This figure depicts the messages that are sent between these components during the hardware demonstrations.}
    \label{fig:turtlebot}
\end{figure}

For the real-world demonstrations, we used a TurtleBot kit, which includes a Kobuki mobile base, an Orbbec Astra camera, and a Gigabyte laptop computer (Figure \ref{fig:turtlebot}). To control the TurtleBot, we designed a controller that receives odometry measurements from the Kobuki mobile base and receives RGB color and depth images from the Orbbec Astra camera. The controller then uses this input to determine the position, velocity, angle, and radius of the robot and surrounding pedestrians. This information about the state of the environment is fed into our socially-aware RL policy to determine the desired velocity of the robot. The TurtleBot controller receives this velocity and sends the appropriate command to the mobile base to move the robot. This process is carried out using the Robot Operating System (ROS) and is summarized in Figure \ref{fig:turtlebot}.

\subsubsection{Pedestrian Detection}

\begin{figure}[h!]
    \centering
    \captionsetup{width=\linewidth}
    \includegraphics[width=\textwidth]{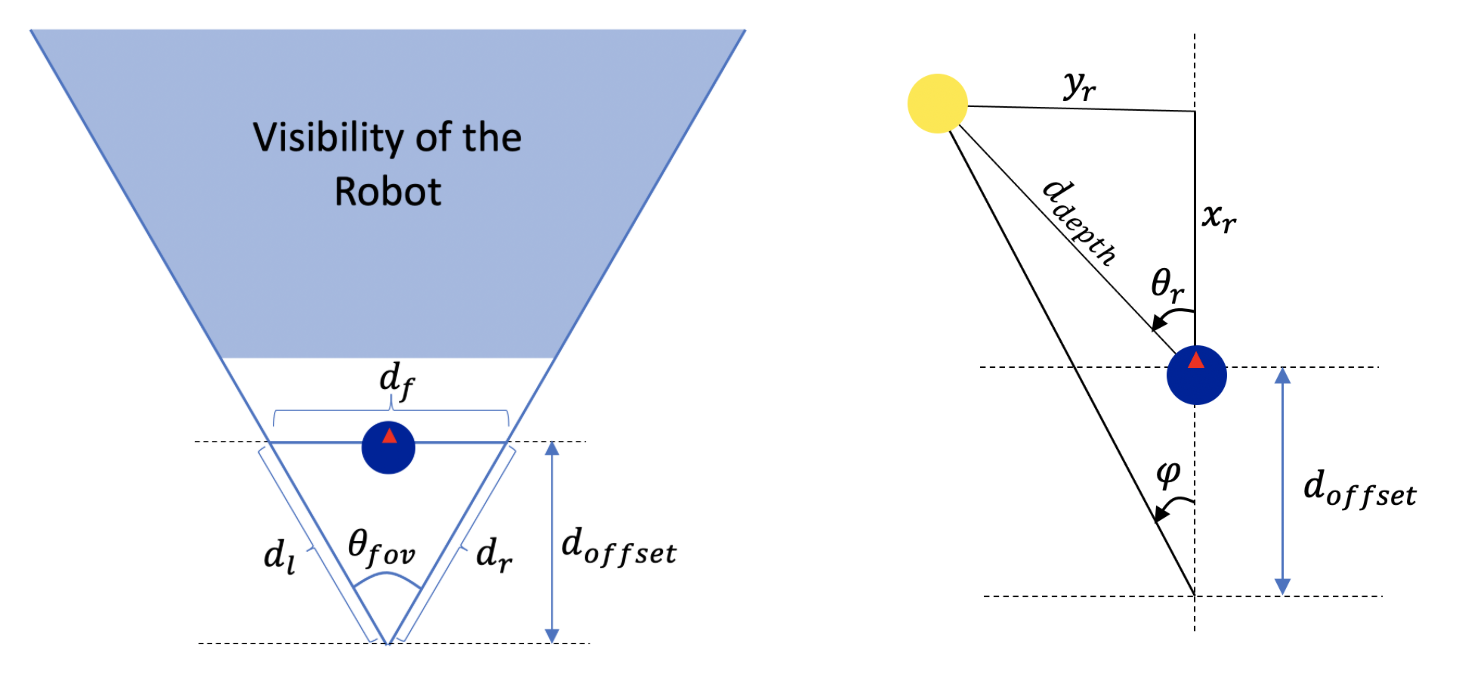}
    \caption{(Left) To calibrate the camera, the field of view (light blue region) of the robot (dark blue circle) is used to determine the offset distance, $d_{offset}$, and field of view angle, $\theta_{fov}$. (Right) The variables used to compute the position of a pedestrian (yellow circle) with respect to the robot (blue circle). In this diagram, $d_{offset}$ is again the offset distance determined during camera calibration, $\phi$ is the camera angle, $d_{depth}$ is the distance of the object from the robot, $\theta_r$ is the angle between the object and robot, and $(x_r,y_r)$ is the position of the object with respect to the robot.}
    \label{fig:detection}
\end{figure}

Before performing the real-world demonstrations, the camera needed to be calibrated to accurately detect pedestrians within the robot's field of view. We determined the offset distance ($d_{offset}$) and field of view angle ($\theta_{fov}$) as shown on the left in Figure \ref{fig:detection}.

To determine the position of a pedestrian in the robot's field of view, we use the YOLO object detection system \citep{yolo}. This system passes an RGB color image through a neural network to generate bounding boxes around each person in the image. We use these bounding boxes to determine the position of the pedestrians relative to the robot. To do so, we first determine the horizontal position of the pedestrian. Let $(x_0,y_0)$ be the bottom left corner of a bounding box in pixels and $(x_1,y_1)$ be the top right corner. The horizontal position of the center of the bounding box is then given by
\[
x_c = x_0 + \frac{x_1 - x_0}{2}.
\]

We use the horizontal pixel position of the pedestrian, $x_c$, and the field of view angle, $\theta_{fov}$, to compute the camera angle of the pedestrian ($\phi$ in Figure \ref{fig:detection}). If the image has a width of $W$ pixels, this angle is given by
\[
\phi = \left(\frac{x_c}{W}
\right)\theta_{fov} - \frac{\theta_{fov}}{2}.
\]

We then use the bounding box obtained from YOLO in combination with the depth image to determine the distance of the object from the robot ($d_{depth}$ in Figure \ref{fig:detection}). More specifically, we use the pixels in the RGB image corresponding to bounding box of the obstacle to determine the depth at each associated position in the depth image. Ignoring regions where the view of the obstacle is obstructed and where the depth sensor cannot get a good measurement, we compute the median of the depth measurements of the pedestrian to estimate its distance from the robot. This distance is then adjusted to account for inaccuracies in the depth camera, which depend on the angle $\phi$. The next step is to find the angle of the obstacle with respect to the robot ($\theta_r$ in Figure \ref{fig:detection}), using the following relationship:
\[
\theta_r = \phi + \sin^{-1}\left(\frac{d_{offset}\sin(\phi)}{d_{depth}}\right).
\]

Using the depth measurement and the angle of the obstacle with respect to the robot, we can then determine the position of the pedestrian with respect to the robot. The horizontal position, $x_r$, and vertical position, $y_r$, (shown in Figure \ref{fig:detection}) are computed as
\[
x_r = d_{depth} \cos(\theta_r)
\hspace{5mm}\text{and}\hspace{5mm}
y_r = d_{depth} \sin(\theta_r).
\]

The position and angle of the robot with respect to the world frame are determined from the odometry measurements provided by the mobile base. Let $\bigl(p_x^{(r)},p_y^{(r)}\bigr)$ be the position of the robot with respect to the global origin and $\theta^{(r)}$ be the orientation of the robot. Using these values, the position of the pedestrian with respect to the world frame is given by $(p_x^{(h_i)}, p_y^{(h_i)})$, where
\[
p_x^{(h_i)} = x_r \cos\bigl(\theta^{(r)}\bigr) - y_r \sin\bigl(\theta^{(r)}\bigr) + p_x^{(r)},
\]
\[
p_y^{(h_i)} = x_r \sin\bigl(\theta^{(r)}\bigr) + y_r \cos\bigl(\theta^{(r)}\bigr) + p_y^{(r)}.
\]

\subsubsection{Pedestrian Tracking}

Because the robot has a limited field of view, pedestrians the robot detects are expected to remain nearby after they are no longer visible to the robot. To account for this, we developed a method to track detected pedestrians. Each time the robot detects a new pedestrian, the current position of this person is stored in memory. The next time this person is detected, its change in position over time is used to estimate its velocity. At each time step, the current position and velocity of the pedestrian are used to predict the position of the pedestrian at the next time step. The expected next position of previously seen pedestrians is then used to predict whether a recently detected pedestrian is one of the previously seen pedestrians. If a detected pedestrian is believed to be the same as an already existing pedestrian, that person's position is updated to be the position determined from the current image. The velocity is also updated using the expected position and the position determined by the current image. After pedestrians exit the robot's field of view, they are assumed to continue moving at their last seen velocity, and they remain in memory for a short period of time after they were last detected, depending on the number of time steps the person was visible to the robot.

\subsection{Additional Results \& Analysis}

\subsubsection{Navigation Policies with \ORCAeps{} Pedestrians} \label{appendix:more_results}

\begin{figure*}[h]
\includegraphics[width=\linewidth]{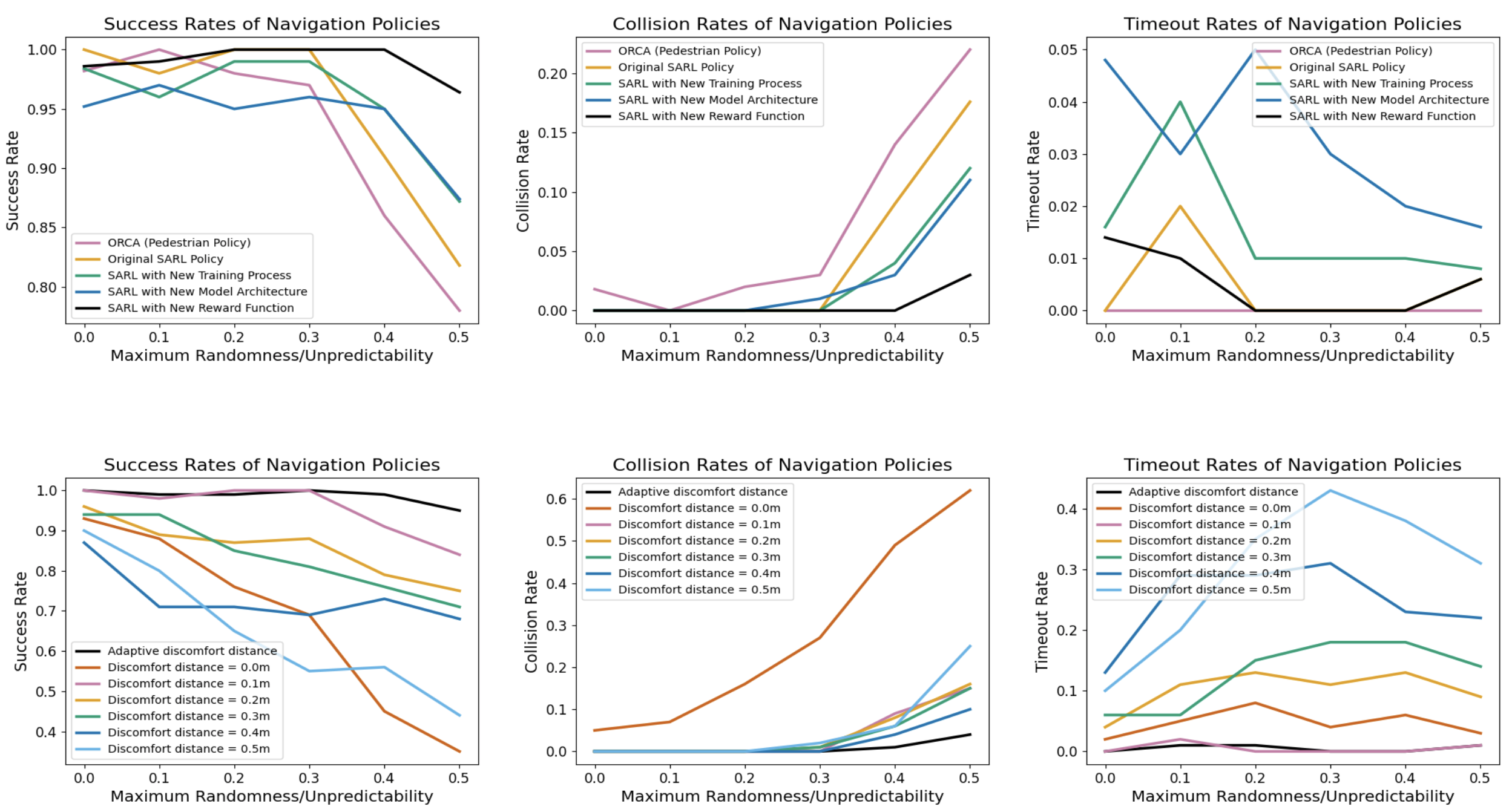}
\centering
\caption{(Top) An ablation study of our uncertainty-aware social navigation RL policy with \ORCAeps{} pedestrians. (Bottom) A comparison of our uncertainty-aware policy with an uncertainty-dependent discomfort distance to RL policies with fixed discomfort distances. (Left) Success rates. (Middle) Collision rates. (Right) Timeout rates.
}
\label{fig:more_plots}
\end{figure*}

Figure \ref{fig:more_plots} shows the success rates, collision rates, and timeout rates for our ablation study in the CrowdSim environment with Noisy ORCA pedestrians (from Section \ref{sec:ablation}). We also expand on the discomfort distance study (from Section \ref{sec:discomfort}) and summarize the results in Figure \ref{fig:more_plots}.

\subsubsection{Parameterization Experiments for the Uncertainty Network} \label{appendix:uncertainty}

\begin{figure*}[h]
     \centering
     \begin{subfigure}[b]{0.24\textwidth}
         \centering
         \includegraphics[width=\textwidth]{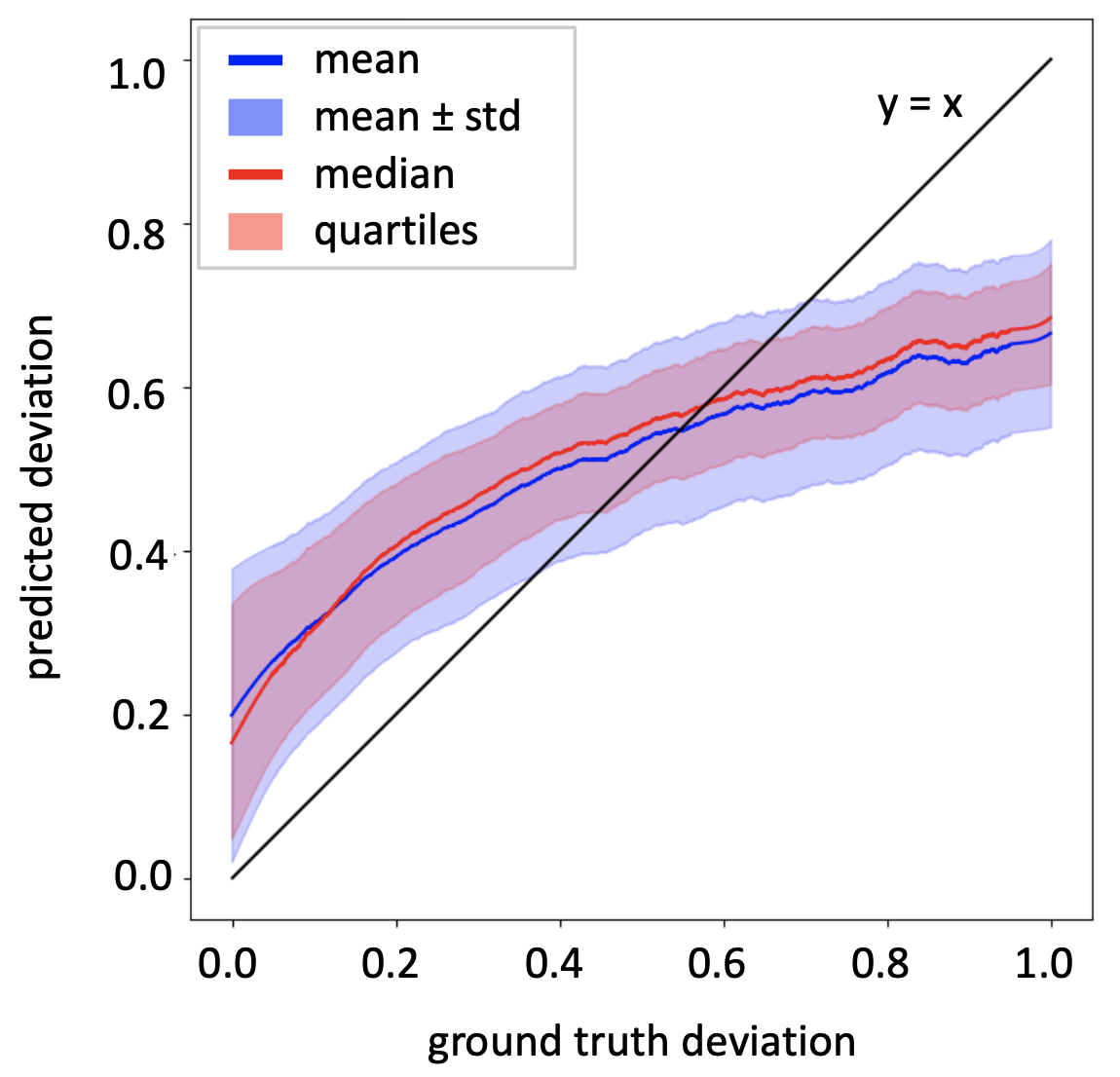}
         \caption{1 time step}
     \end{subfigure}
     \hfill
     \begin{subfigure}[b]{0.24\textwidth}
         \centering
         \includegraphics[width=\textwidth]{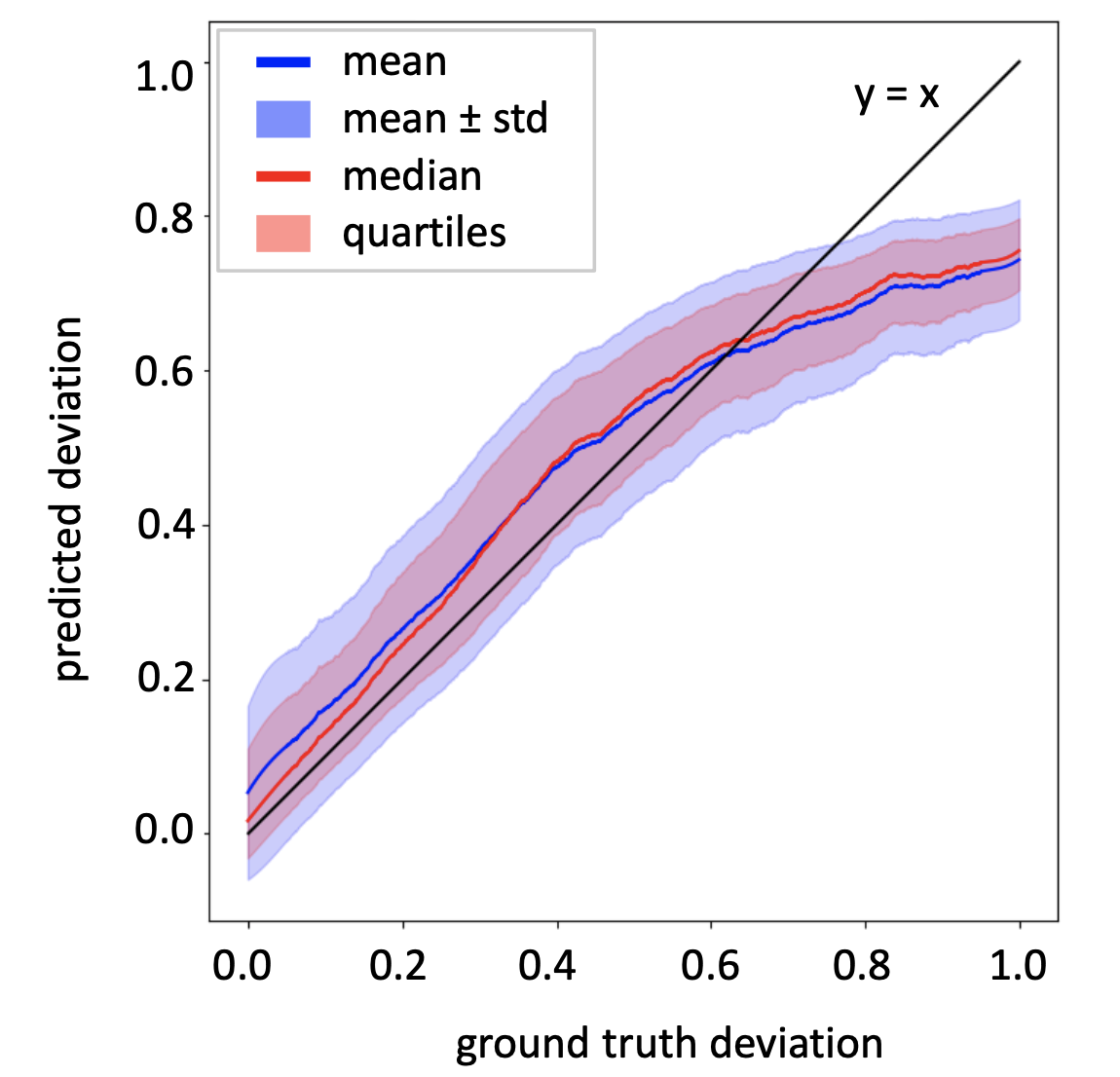}
         \caption{5 steps}
     \end{subfigure}
     \hfill
     \begin{subfigure}[b]{0.24\textwidth}
         \centering
         \includegraphics[width=\textwidth]{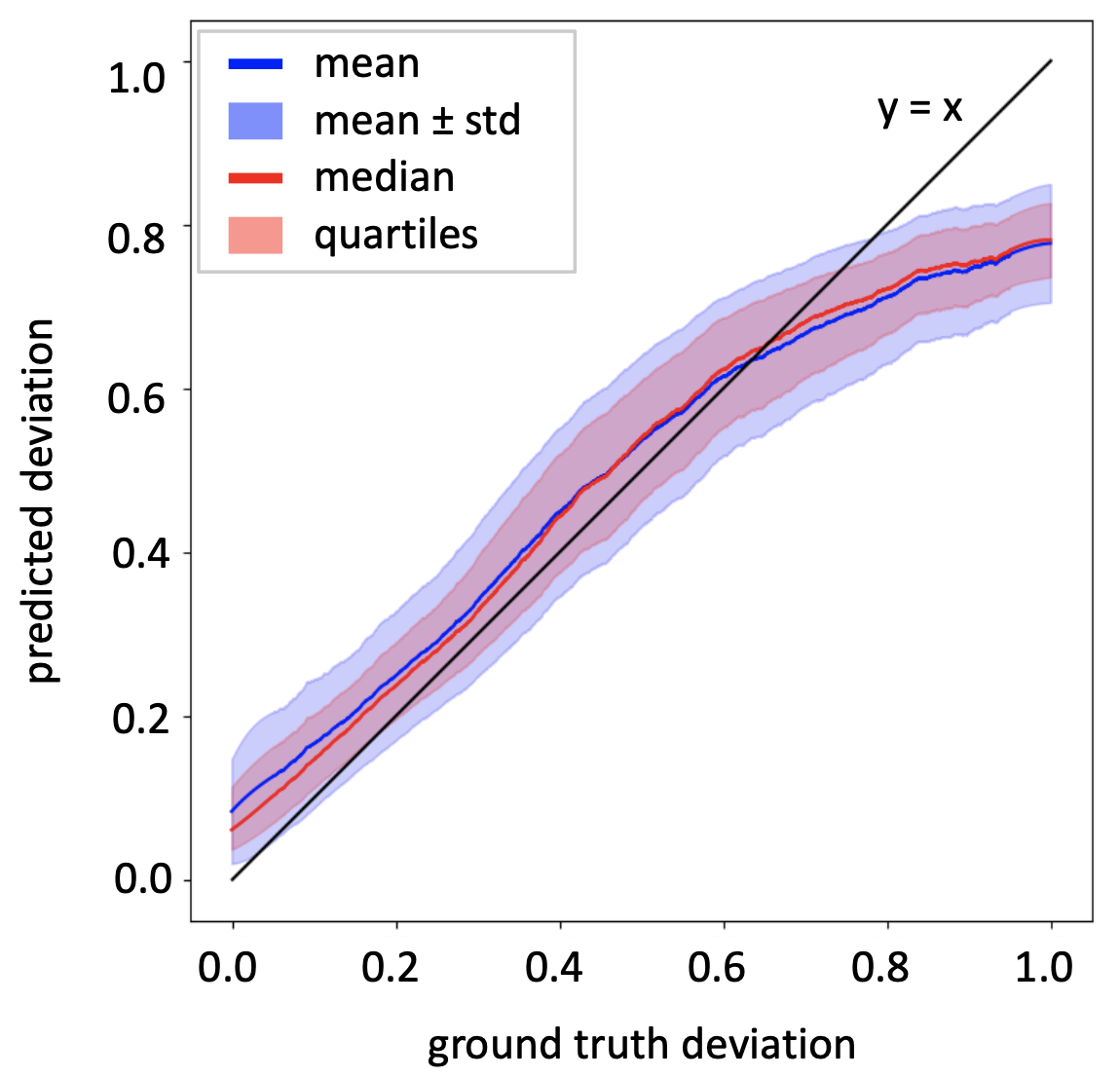}
         \caption{10 steps}
     \end{subfigure}
     \hfill
     \begin{subfigure}[b]{0.24\textwidth}
         \centering
         \includegraphics[width=\textwidth]{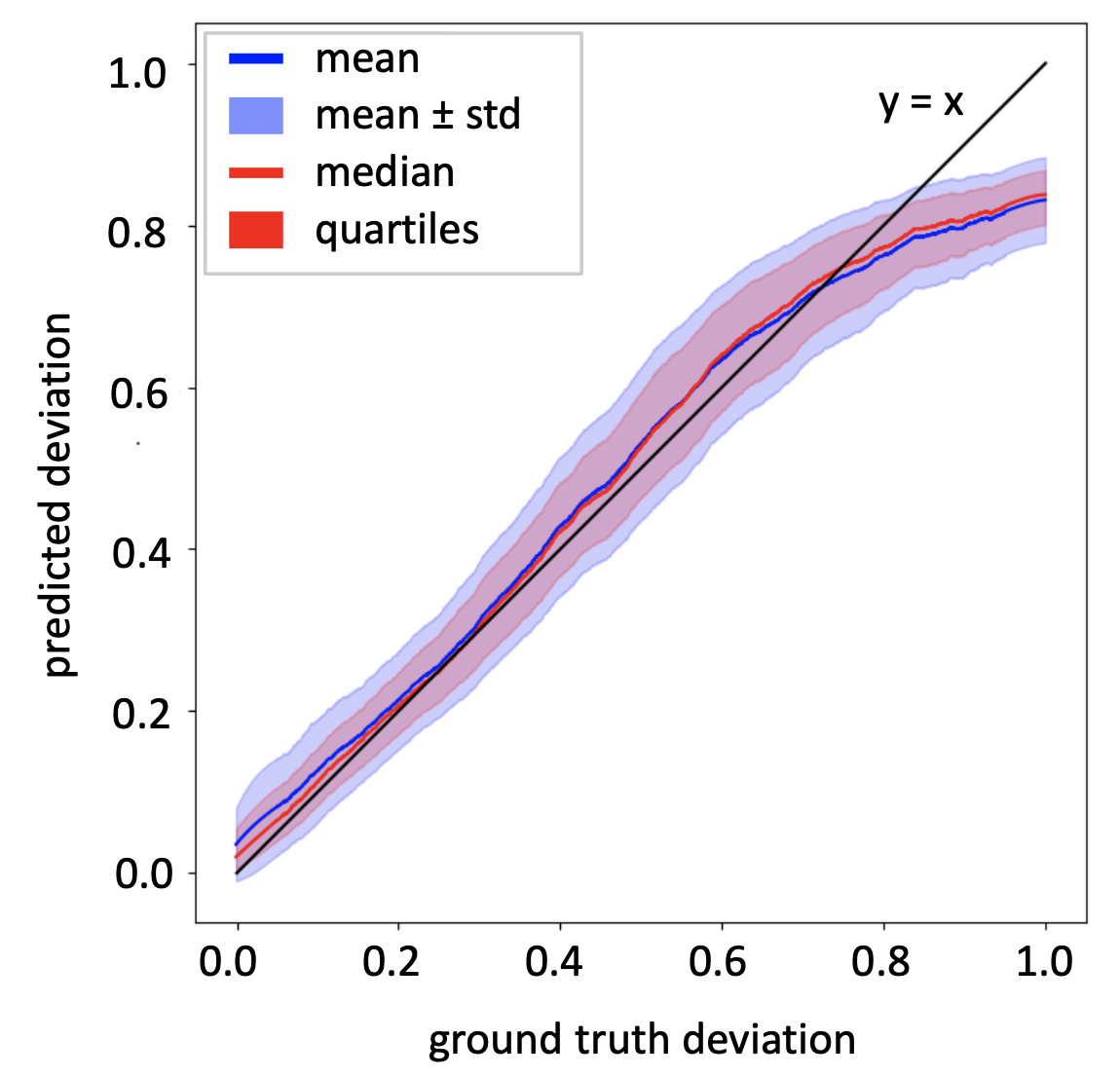}
         \caption{20 steps}
     \end{subfigure}
        \caption{Estimated deviation values for a progressively larger number of time steps used for prediction. Ground truth deviation ($\rho$) values are along the horizontal axis, and estimated values ($\hat{\rho}$) are along the vertical axis. Mean and standard deviation is plotted in blue. Median and quartiles are plotted in red. Accuracy and precision improve as more time passes.}
        \label{fig:final_validation}
\end{figure*}

To design the uncertainty estimation network (MLP$_1$ in Figure \ref{fig:nn}), we considered a number of different model architectures (Section \ref{appendix:model}) and input preprocessing methods (Section \ref{appendix:features}). We found that increasing the number of time steps in the input has the greatest impact on model accuracy. In general, as we increased the number of time steps, the predicted uncertainty values more closely aligned with the ground-truth uncertainty values, as we show in Figure \ref{fig:final_validation}. We find that at 20 time steps, the predicted $\epsilon$ values are reasonably accurate with a standard deviation of around 0.1, and the performance does not significantly improve with additional time steps.

\end{document}